\documentclass[conference]{IEEEtran}
\IEEEoverridecommandlockouts
\usepackage[noadjust]{cite}
\usepackage[pdftex]{graphicx}
\graphicspath{{Images/}}
\usepackage{color}
\usepackage{amsmath}
\ifCLASSOPTIONcompsoc
 \usepackage[caption=false,font=normalsize,labelfont=sf,textfont=sf]{subfig}
\else
 \usepackage[caption=false,font=footnotesize]{subfig}
\fi

\begin{document}
\bstctlcite{MyBSTcontrol}

\title{Seafloor Classification based on an AUV Based Sub-bottom Acoustic Probe Data for Mn-crust survey}

\author{\IEEEauthorblockN{Umesh Neettiyath$^{1}$, Harumi Sugimatsu$^{1}$, Blair Thornton$^{2,1}$}
\IEEEauthorblockA{$^{1}$Institute of Industrial Science, The University of Tokyo, Komaba, 153-8505, Japan \\(E-mail: umesh@iis.u-tokyo.ac.jp)\\
$^{2}$Southampton Marine and Maritime Institute, University of Southampton, Southampton SO16 7QF, UK\\}}

\maketitle


\IEEEpubid{\begin{minipage}{\textwidth}\ \\[12pt]
This is a pre-print of the manuscript accepted for publication at IEEE/MTS Oceans Conference in GulfCoast, 2023. The final paper will be published in IEEE explore after the conference [2023/09/25].
\end{minipage}}

\IEEEpubidadjcol

\begin{abstract}
The possibility of automatically classifying high frequency sub-bottom acoustic reflections collected from an Autonomous Underwater Robot is investigated in this paper. In field surveys of Cobalt-rich Manganese Crusts (Mn-crusts), existing methods relies on visual confirmation of seafloor from images and thickness measurements using the sub-bottom probe. Using these visual classification results as ground truth, an autoencoder is trained to extract latent features from bundled acoustic reflections. A Support Vector Machine classifier is then trained to classify the latent space to idetify seafloor classes. Results from data collected from seafloor at 1500m deep regions of Mn-crust showed an accuracy of about 70\%.
\end{abstract}

\IEEEpubidadjcol

\IEEEpeerreviewmaketitle

\section{Introduction}
Deep sea mineral deposits, in particular, Cobalt-rich manganese crusts (Mn-crust) are gaining popularity as a potential source of valuable minerals including Cobalt, Nickel, and rare earth elements. These are formed mainly on the slopes and shoulders of seamounts ranging from 800m to 2400m~\cite{unep_crust1, Usui2016}. Stakeholders have attempted different methods of surveying such as core drills for detailed analysis, towed camera surveys for visual inspection and ship based multibeam backscatter studies for large area surveys. While sampling gives high accuracy thickness measurements with very low spatial resolution, multibeam and camera studies cannot measure thickness values. Thickness and coverage are important results necessary to estimate the volume of Mn-crust; and an integrated approach is necessary to accurately estimate the resource potential.

\begin{figure}[!h]
\includegraphics[trim = 0mm 10mm 0mm 0mm, clip, width=\columnwidth]{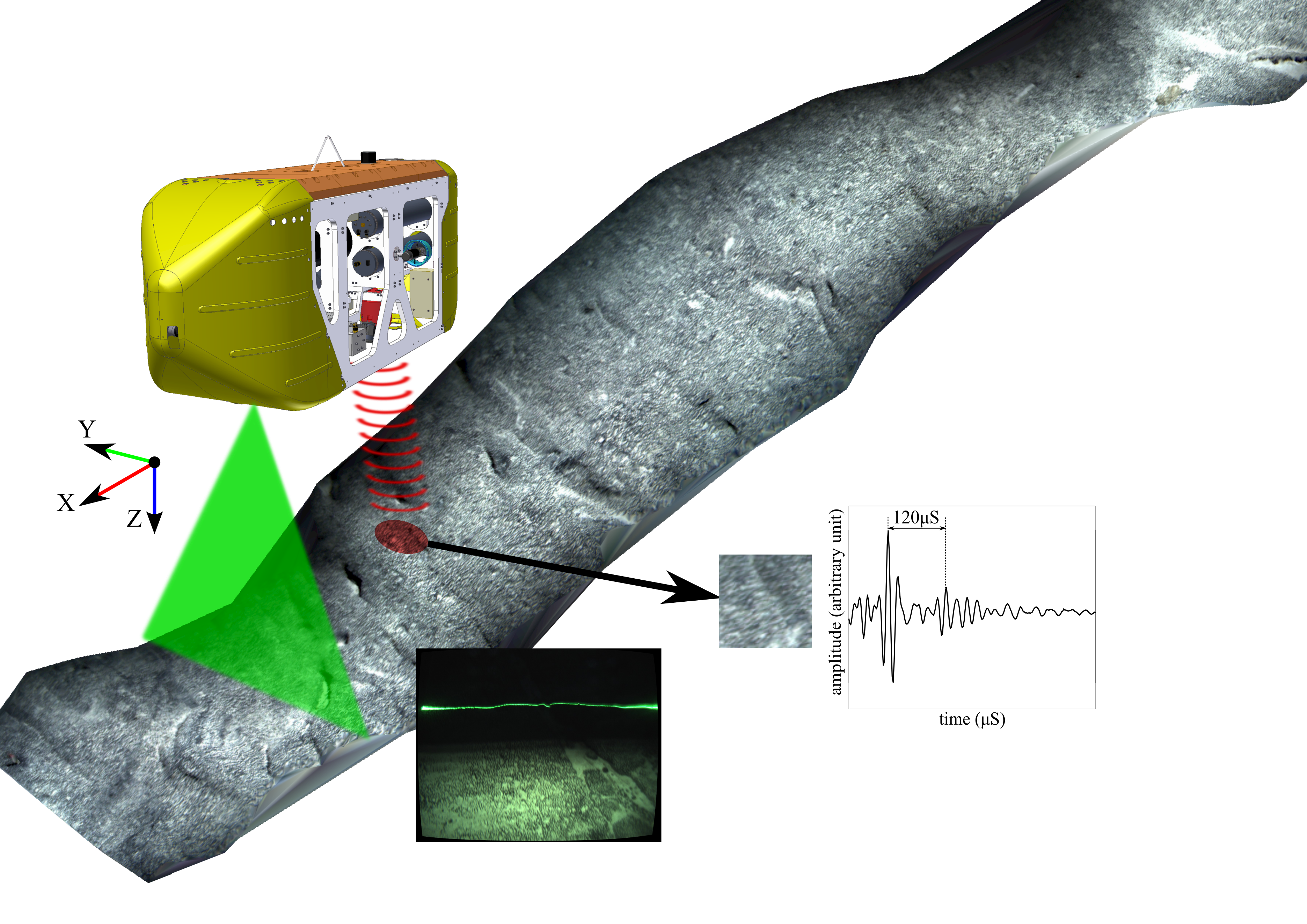}
\caption{Illustration of the mapping scenario using AUV Boss-A. The robot surveys the seafloor using an acoustic subbottom probe and a visual 3D mapping system. Examples of images collected and acoustic reflections from Mn-crust areas are shown.}
\label{figOverview}
\end{figure}

Towards this goal, an acoustic probe for in-situ contactless thickness measurement, and an Autonomous Underwater Vehicle (AUV), called Boss-A, to conduct continuous surveys were developed by the Institute of Industrial Science of the University of Tokyo~\cite{Thornton2013, Nishita2016}, an Illustration of which is given in Fig.~\ref{figOverview}. A method for calculating the volumetric distribution of Mn-crust was developed by combining the thickness measured from the acoustic probe with the coverage of exposed Mn-crust calculated from 3D visual maps~\cite{umesh2020}. The system is being successfully deployed for Mn-crust volumetric surveys regularly.

This was achieved by classifying the 3D colour point cloud of the seafloor generated using a Support Vector Machine classifier (SVM)~\cite{Umesh2015}. Typically occurring seafloor types - Mn-crust, nodules, and sediment deposits were the target classes, as shown in Fig.~\ref{figExample}. In areas detected as Mn-crust, a thickness was calculated from the acoustic reflections. In the existing volumetric estimation method, although the seafloor characterization was done from the visual data~\cite{umesh2020}, it was observed that the seafloor reflections from different areas appear different. The proposed paper intends to create a new classifier for classifying acoustic data without relying on visual information. Also, it is possible that features not visible in the visual images can be observed in sub-bottom scans, leading to further insights during seafloor characterization.

Rest of this paper is organized as follows. A description of the acoustic and visual systems are described in section II. Section III details the algorithms, and the data processing methods. Results from field surveys and discussion are gievn in section IV. Section V contains the concluding remarks and future directions.

\section{System Overview}
\label{secSystem}

The AUV Boss-A system consist of visual and acoustic survey subsystems as shown in Fig~\ref{figOverview}. These systems are supported by the navigational system on the AUV - Doppler Velocity Log, AHRS systems, and depth sensor. The specifications of the systems are given in Table~\ref{tabSystem}.

\begin{table}
\caption{Specifications of the systems used in the survey}
\begin{center}
\begin{tabular}{ |p{4cm} p{4cm}| } 
\hline
\textbf{Visual mapping system} & \\
\hline
Camera resolution & $1328 \times 1048$\\
Horizontal opening angle & $65^{\circ}$\\ 
Vertical opening angle & $53^{\circ}$ \\ 
Laser to camera baseline & 1.22 m \\
Frame rate & 15 fps \\
\hline
\hline
\textbf{Acoustic Probe} & \\
\hline
Frequency & 2 MHz (carrier), 200kHz (signal) \\ 
-3 dB footprint & $<2$  cm (dynamic focusing)\\ 
Ping rate & 20 Hz \\ 
Sampling rate & 2 MHz\\
\hline
\end{tabular}
\end{center}
\label{tabSystem}
\end{table}

\subsection{Visual mapping system}
The visual mapping system is based on the light-sectioning based seafloor 3D mapping system developed in~\cite{Bodenmann2017}. It consists of a sheet laser, LEDs for illumination, and a camera which records the laser projection on the seafloor. From each recorded image, the laser deformation is calculated and used to calculate the seafloor bathymetry. From the LED illuminated regions, RGB colour values of each point on the seafloor is recognised and combined with the bathymetry information to generate 3D color reconstructions of the seafloor. The resolution of the generated 3D map is \texttildelow 1.4\,mm.

\subsection{Acoustic sub-bottom probe}
\begin{figure}[t]
\includegraphics[trim = 0mm 0mm 0mm 0mm, clip, width=\columnwidth]{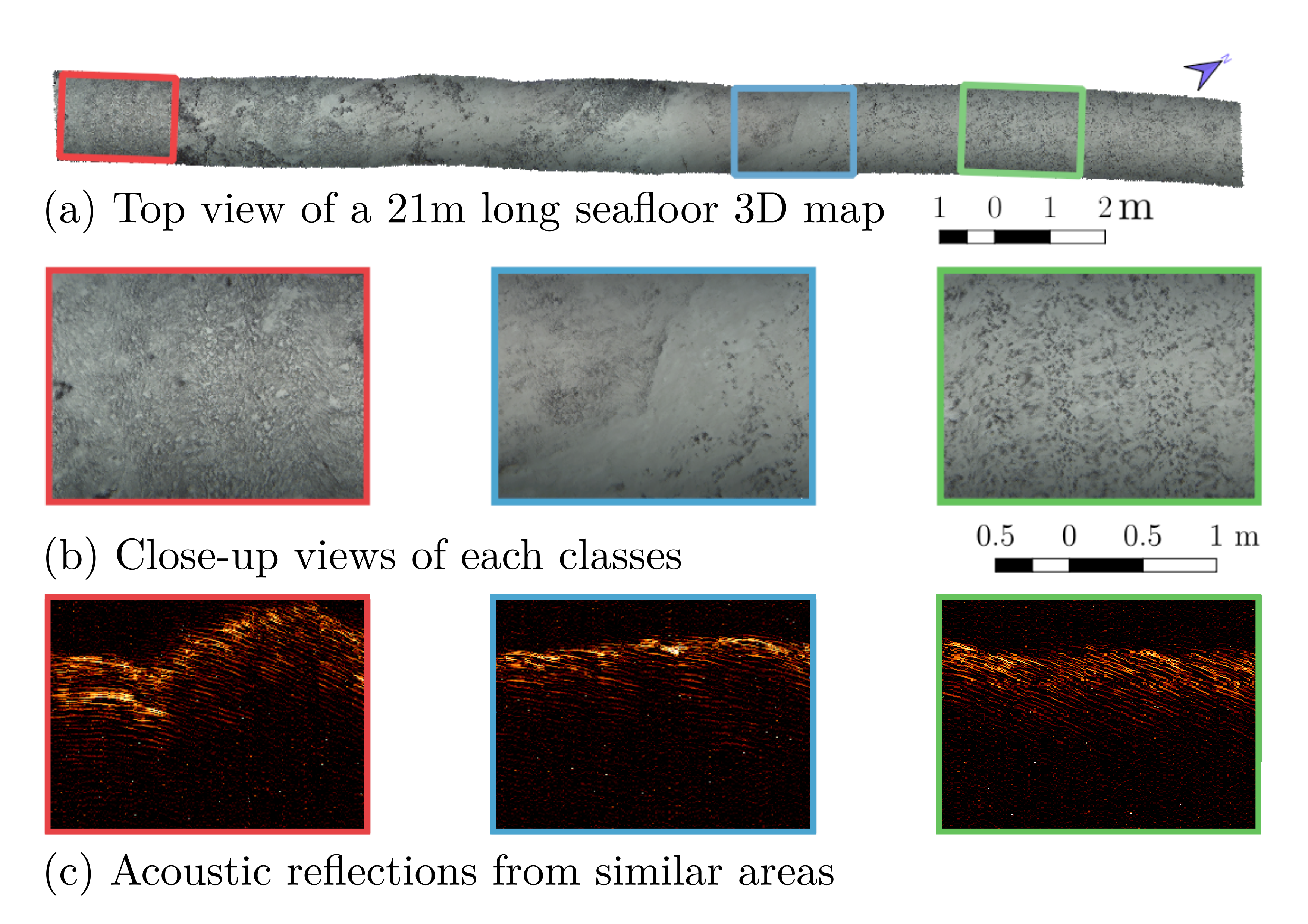}
\caption{Examples of typical seafloor in Takuyo Daigo seamount at ~1500m depth. Flat Mn-crust (red), sediments (blue), and Nodules (green) have distinct acoustic sub-bottom and visual footprints.}
\label{figExample}
\end{figure}

The acoustic probe is a parametric subsurface sonar that records subsurface reflections of the seafloor for estimating the Mn-crust thickness~\cite{Thornton2013}. The probe consists of a five-channel annular array of 2-MHz piezoelectric transducers for transmission and a 200-kHz piezoelectric transducer to record reflections. It is dynamically focused on the seafloor at ranges from 0.5 to 2.5 m. The recorded signal typically consists of reflections from the top of the seafloor and the bottom of the Mn-crust layer in crust covered areas. The thickness can then be calculated by multiplying the time delay between the reflections with the velocity of sound in Mn-crust. In sediment and nodule covered areas, no secondary reflection can be seen, with the exception of buried Mn-crust layers as studied in~\cite{neettiyath_automatic_2022}. As shown in Fig.~\ref{figExample}(c), while only Mn-crust areas have secondary reflections, the primary reflections are also characteristic of the seafloor type.

\section{Algorithm and Processing Workflow}
\label{secField}

\begin{figure*}[!h]
\includegraphics[trim = 0mm 0mm 0mm 0mm, clip, width=0.9\textwidth]{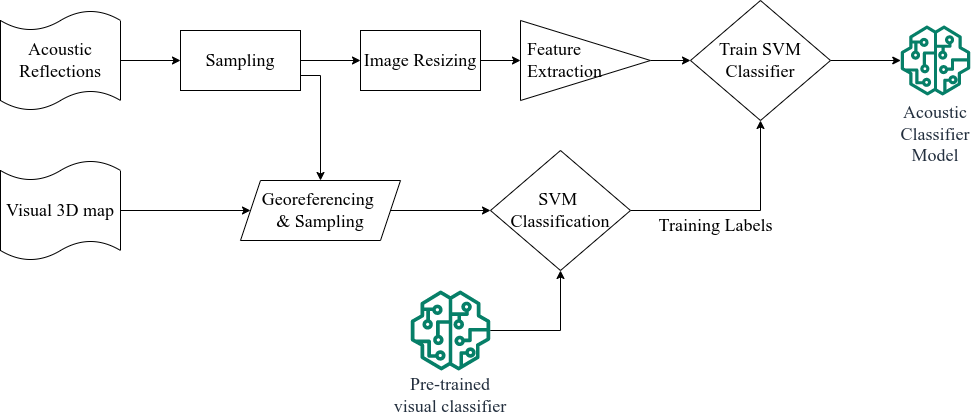}
\caption{Flowchart of the proposed classification method. Using previously developed classifier for visual 3D data as ground truth, a classifier for acoustic subbottom reflections from co-located areas are trained.}
\label{figAlgo}
\end{figure*}

A flowchart of the  proposed algorithm is given in Fig.~\ref{figAlgo}. The visual 3D colour reconstruction of the seafloor is divided into 10cm x 10cm cells and each cell is classified into Mn-crust, sediment or nodules by an SVM classifier, developed in \cite{Umesh2015}. Each sample is georeferenced to identify cells in which acoustic sub-surace reflections are also made. Since the generated 3D map is \texttildelow 2\,m in width, whereas the acoustic data is collected along a line in the middle, only a fraction of the cells would be selected.

\begin{figure}[h]
\centering
\includegraphics[trim = 0mm 10mm 0mm 0mm, clip, width=0.7\columnwidth]{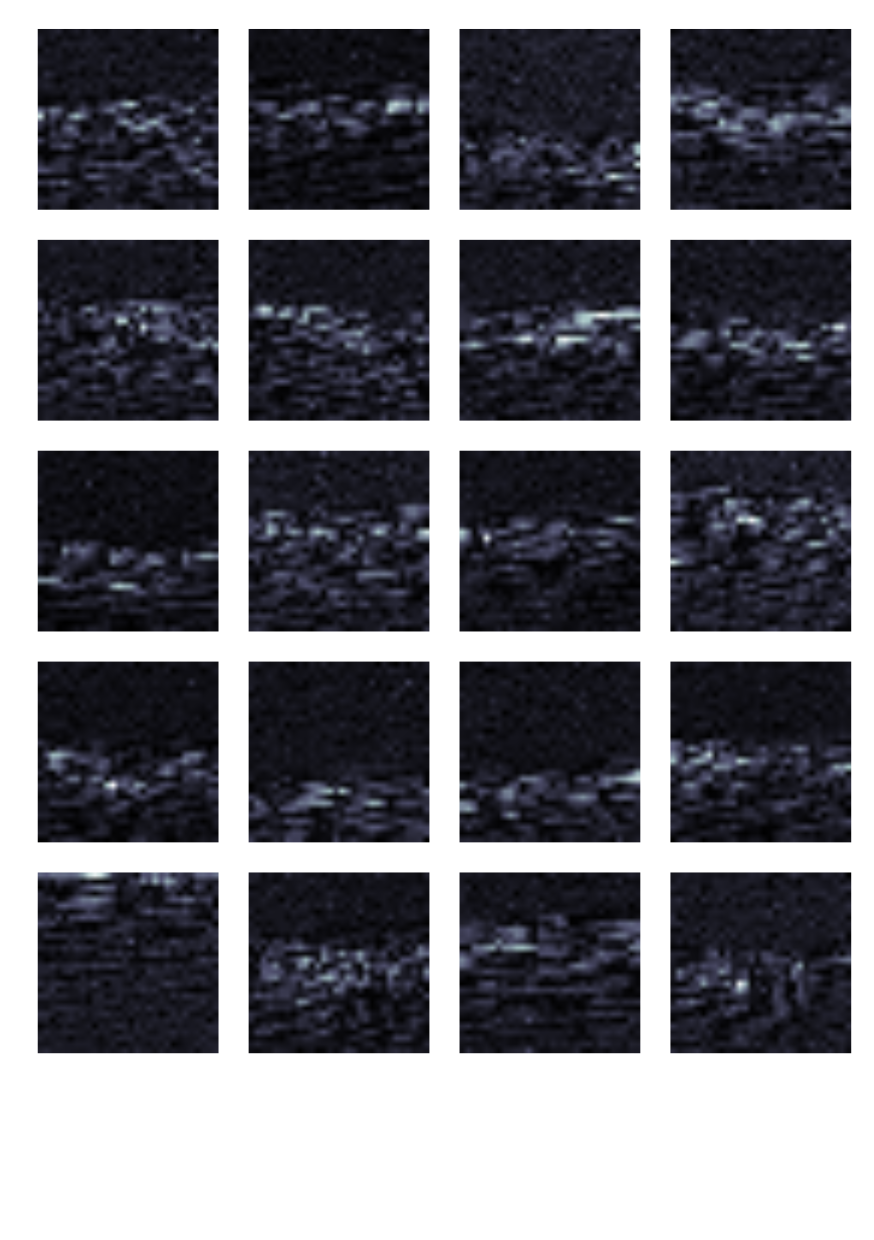}
\caption{Acoustic tiles used for feature extraction}
\label{figTiles}
\end{figure}

The acoustic reflections collected are stacked into time-series images, sections of which are shown in Fig.~\ref{figExample}(c). This image is sampled to select the reflections around the point of incidence of the acoustic pulse on the seafloor. The size of each sample(called an acoustic tile) is 30x30 pixels. By using speed of sound in Mn-crust (2932~\textpm~179\,m/s), the selected region corresponds to an approximate physical size of 44\,mm in the vertical direction, which is large enough to capture the top reflection, while excluding the unwanted reflections. In the horizontal direction, the acoustic tile would have a size of 15\,cm assuming a constant AUV velocity. Fig.~\ref{figTiles} shows some of the acoustic tiles generated. The Mn-crust thickness values are also calculated from the acoustic data using the method demonstrated in~\cite{Umesh2017}, but is not included in the current paper.

The acoustic tiles generated are used as the input into a feature extractor in order to generate a minimal vector which can accurately represent the data. This was achieved using an autoencoder based on Alexnet~\cite{Yamada2020}. Alexnet is a deeply stacked artificial neural network with convolutional layers and has been used in a variety of image data extraction applications~\cite{Cheng2016, Yamada2021}. In ~\cite{Yamada2020}, Alexnet was combined with an inverted version of Alexnet in order to reconstruct the images from the latent vectors. In addition, physical proximity of the samples is considered into a loss function, bearing the idea that images (in this case, acoustic tiles) collected from nearby areas are more likely to be similar. An overall loss function is calculated by combining reconstruction loss with proximity loss. The autoencoder is then trained to minimize the loss during reconstruction of the images.

Once the training is completed, the Alexnet classifier is used to generate latent vectors for each acoustic tile. The extracted feature vectors are classified using a second Support Vector Machine (SVM) classifier. SVM is an unsupervised classification method widely used for image classification and segmentation tasks~\cite{scikit-learn}. It works by generating a hyperplane in an N-dimensional space which maximizes the separation between the nearest training input data samples. The input data consists of a feature vector, which is the latent space representation of generated by the autoencoder in the previous step and training labels, which are generated by classifying the co-located visual 3D maps using the first SVM classifier. The training labels belong to the 3 classes of seafloor typically found in Mn-crust covered seamounts - Mn-crust, sediment and nodules, as shown in Fig.~\ref{figExample}. The training and testing of the classifier using field collected data is done using machine learning workflows and is described in the following section.

\section{Implementation and Results}
\label{secAnalysis}

AUV Boss-A was deployed in field experiments at Takuyo Daigo seamount in the northwestern Pacific Ocean at depths ranging from 1350\,m to 1650\,m~\cite{umesh2020} during the KR16-01 cruise of R/V Kairei of the Japan Agency for Marine Earth Science and Technology. The AUV surveyed the seafloor at a velocity of 0.2\,kn (0.1\,m/s) at a target altitude of 1.5\,m.

Two transects of the robot are selected to study the proposed method, as shown in Table~\ref{tabDatasets}. Non-overlapping samples were selected at nearly uniform intervals to form two datasets. It can be seen that most of the samples are Mn-crust, since the target area in which the surveys were conducted was Mn-crust. Efforts were made to remove overfitting the classifier by normalising the input data.

\begin{table}
 \caption{Two datasets used for training and testing the classifier. 30 percent of the points were randomly selected to prepare the test set.}
\begin{center}
\begin{tabular}{cccccc}
\hline
 & & & No. of samples & \\
Transect & Area ($m^2$) & Mn-crust & Sediment & Nodules & Total\\
\hline
A & 970 & 730 & 189 & 377 & 1296\\
B & 462 & 335 & 231 & 106 & 672\\
\hline
\end{tabular}
\end{center}
\label{tabDatasets}
\end{table}

The two datasets were combined and sampled randomly to generate training and testing datasets. 70\% of the points were assigned to the training set and 30\% of the points were assigned to the test set. Training test was used to train the SVM classifier and the performance of the classifier was examined using the test set.

\begin{figure}[h]
\includegraphics[trim = 0mm 0mm 0mm 0mm, clip, width=\columnwidth]{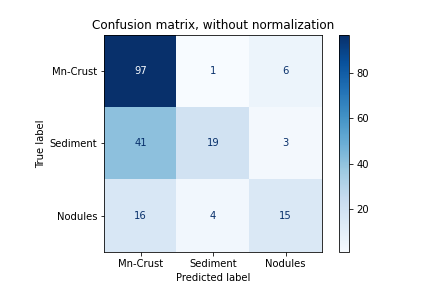}
\caption{Confusion matrix of the classifier on the test set. It can be seen that the prediction is biased towards Mn-crust for all classes.}
\label{figConfMx}
\end{figure}

The results of the test set are shown in Fig.~\ref{figConfMx}. It can be seen that Mn-crust is classified accurately (93\% accuracy), whereas performance is lower in other classes, some of sediment and Nodules were classified wrongly as Mn-crust. While it could be an issue due to the imbalance of the training set, efforts were made to reduce overfitting. Some areas were seen which contained Mn-crusts with a very thin sediment layer of a few millimeters or less. In such areas, while the visual images indicate sediments, acoustic reflections could indicate Mn-crusts. Further analysis is required into these specific areas to fully understand this scenario.

\begin{figure*}[h]
\includegraphics[trim = 0mm 0mm 0mm 0mm, clip, width=0.9\textwidth]{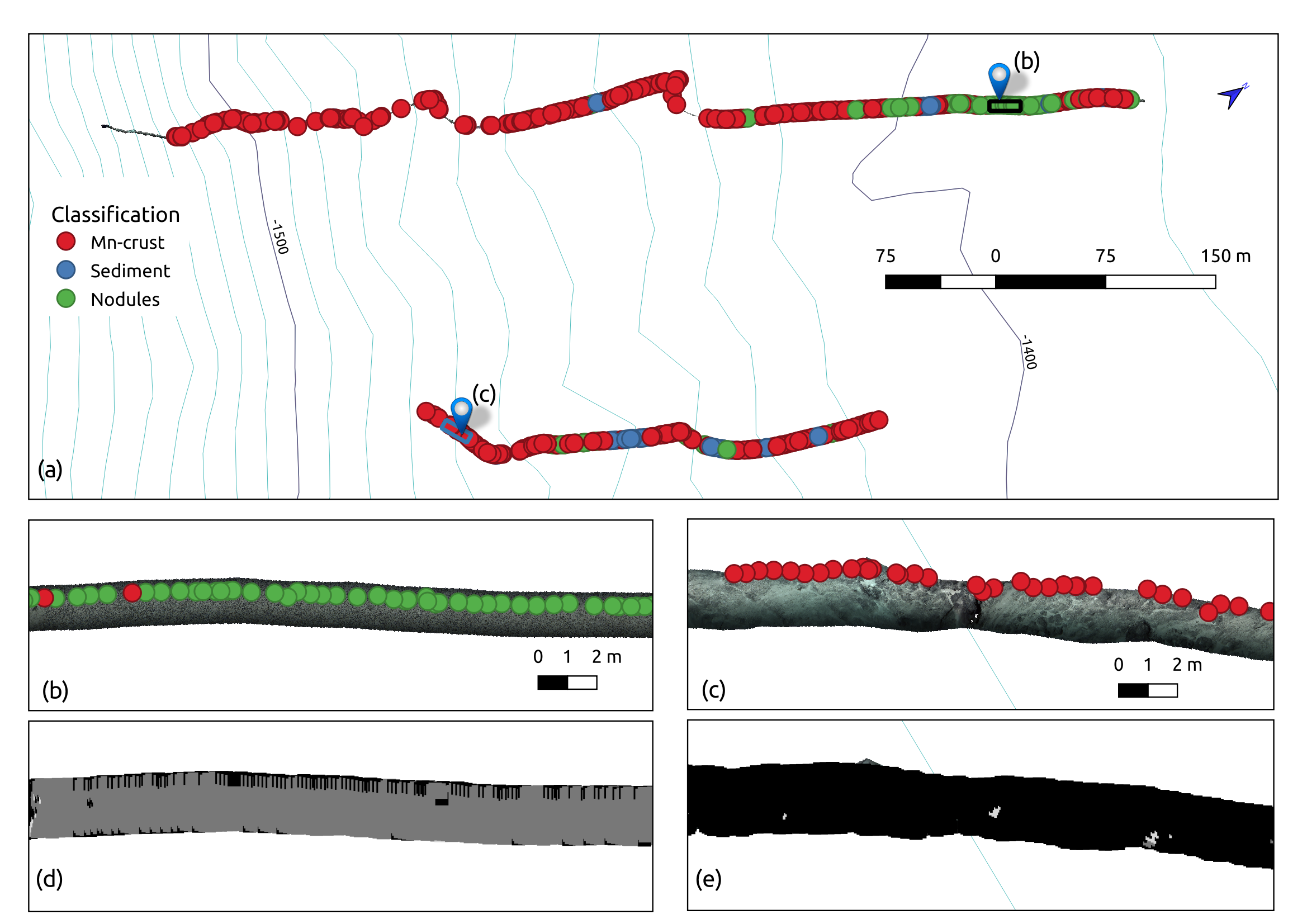}
\caption{Results of classification of two transects surveyed by the AUV at ~1500\,m depth. Most of the area is covered in Mn-crust, with occasional sections of sediment and nodules. (a) Overall results. Two selected sections containing (b)sediment and (c)Mn-crust are also shown. (d) and (e) shows the visual classification results for the sections (b) and (c) respectively.}
\label{figAllDive}
\end{figure*}

The classification results of the complete dataset is shown in Fig.~\ref{figAllDive}. The results are overlaid on the top view of the 3D reconstruction for visual confirmation reference. Two selected areas are shown below and their locations are indicated in the overall results. In Fig.~\ref{figAllDive}(b), a nodule covered area is shown. It can be seen that most of the area is classified accurately, except for two points towards the left. The classification results from the visual 3D classification using the existing classifier is shown in Fig.~\ref{figAllDive}(d), where some misclassification of nodules as sediments is visible in the upper portions. This is caused by the limitations of the colour correction algorithm as part of the 3D map generation. This could cause some of the training labels to be incorrect, creating classification errors for nodule tiles. Efforts to correct this are underway. Fig.~\ref{figAllDive}(c) and (d) shows a Mn-crust covered section. The left side of the section has flat Mn-crusts whereas towards the middle, pillowy Mn-crust is visible. In the crevices between the Mn-crusts, sediment deposits are visible. This section has been accurately classified in both visual and acoustic classification schemes. Since different topographies of Mn-crust are visible, the impact of this in the acoustic reflections can be studied further.

If the acoustic probe can reliably characterize Mn-crust covered seafloors and calculate thickness measurements, it is possible to conduct surveys without a visual mapping system, thereby simplifying the systems and reducing the payload requirements. Another potential research direction is identifying the substrate types using these methods. Since substrates are buried under several centimeters of Mn-crust, visual or sonar surveys cannot provide information regarding them; at present, sampling is the only available method.

The possibility of using a semi-supervised classification on the extracted feature vector is a potential future research theme. The aim of this will to detect different sub-classes within each class - such as different surface topographies of Mn-crust deposits.  It is useful in cross-validating visual classification schemes and for error correction. It can be used when visual data is unreliable or not available. Further applications in scientific studies into Mn-crust topography and formations are possible.

\section{Conclusion}
\label{secConclusion}
%
This paper proposed a method for automatically classifying the seafloor sub-surface reflections collected by an acoustic sub-bottom sonar probe. The acoustic data is bundled into image tiles around the seafloor incidence point, latent features were extracted using an autoencoder, and was classified into Mn-crust, sediment or nodules. Using a field collected dataset from an AUV at ~1500m depth,the method was validated and found to have an accuracy of 65\% overall and 93\% for Mn-crusts. The method can assist the volumetric estimations of Mn-crust deposits, by cross-validating visual classifier results and reduce errors in areas such as thin sediment covered Mn-crusts. It also introduces the possibility to simplify the survey setup and improve scalability by eliminating the 3D mapping system; albeit losing some useful data.

\section*{Acknowledgment}
The authors would like to thank the team members who participated in the data collection - Yuya Nishida, Kazunori Nagano, Tetsu Koike, and the R/V Kairei crew, Japan Agency for Marine-Earth Science and Technology (JAMSTEC) during the KR16-01 cruise. The data used in this work was collcted during surveys supported by the Japanese Ministry of Education under the Program for the Development of Fundamental Tools for the Utilization of Marine Resources.

\IEEEtriggeratref{11}

\bibliographystyle{IEEEtran}
\bibliography{Oceans2023_GulfCoast,disabl_url}

\begin{thebibliography}{10}
\providecommand{\url}[1]{#1}
\csname url@samestyle\endcsname
\providecommand{\newblock}{\relax}
\providecommand{\bibinfo}[2]{#2}
\providecommand{\BIBentrySTDinterwordspacing}{\spaceskip=0pt\relax}
\providecommand{\BIBentryALTinterwordstretchfactor}{4}
\providecommand{\BIBentryALTinterwordspacing}{\spaceskip=\fontdimen2\font plus
\BIBentryALTinterwordstretchfactor\fontdimen3\font minus
  \fontdimen4\font\relax}
\providecommand{\BIBforeignlanguage}[2]{{%
\expandafter\ifx\csname l@#1\endcsname\relax
\typeout{** WARNING: IEEEtran.bst: No hyphenation pattern has been}%
\typeout{** loaded for the language `#1'. Using the pattern for}%
\typeout{** the default language instead.}%
\else
\language=\csname l@#1\endcsname
\fi
#2}}
\providecommand{\BIBdecl}{\relax}
\BIBdecl

\bibitem{unep_crust1}
M.~R. Clark, R.~Heydon, J.~R. Hein, S.~Petersen, A.~Rowden, S.~Smith, E.~Baker,
  and Y.~Beaudoin, \emph{Deep {Sea} {Minerals}: {Cobalt}-rich {Ferromanganese}
  {Crusts}, a physical, biological, environmental, and technical review},
  E.~Baker and Y.~Beaudoin, Eds.\hskip 1em plus 0.5em minus 0.4em\relax
  Secretariat of the Pacific Community, 2013, publication Title: Deep Sea
  Minerals.

\bibitem{Usui2016}
A.~Usui, K.~Nishi, H.~Sato, Y.~Nakasato, B.~Thornton, T.~Kashiwabara,
  A.~Tokumaru, A.~Sakaguchi, K.~Yamaoka, S.~Kato, S.~Nitahara, K.~Suzuki,
  K.~Iijima, and T.~Urabe, ``Continuous growth of hydrogenetic ferromanganese
  crusts since 17 {Myr} ago on {Takuyo}-{Daigo} {Seamount}, {NW} {Pacific}, at
  water depths of 800–5500 m,'' \emph{Ore Geology Reviews}, vol.~87, pp.
  71--87, Jul. 2017, publisher: Elsevier B.V.

\bibitem{Thornton2013}
B.~Thornton, A.~Asada, A.~Bodenmann, M.~Sangekar, and T.~Ura, ``Instruments and
  methods for acoustic and visual survey of manganese crusts,'' \emph{IEEE
  Journal of Oceanic Engineering}, vol.~38, no.~1, pp. 186--203, Jan. 2013,
  iSBN: 0364-9059 VO - 38.

\bibitem{Nishita2016}
Y.~Nishida, K.~Nagahashi, T.~Sato, A.~Bodenmann, B.~Thornton, A.~Asada, and
  T.~Ura, ``Autonomous {Underwater} {Vehicle} “{BOSS}-{A}” for {Acoustic}
  and {Visual} {Survey} of {Manganese} {Crusts},'' \emph{Journal of Robotics
  and Mechatronics}, vol.~28, no.~1, pp. 91--94, Feb. 2016.

\bibitem{umesh2020}
U.~Neettiyath, B.~Thornton, M.~Sangekar, Y.~Nishida, K.~Ishii, A.~Bodenmann,
  T.~Sato, T.~Ura, and A.~Asada, ``Deep-{Sea} {Robotic} {Survey} and {Data}
  {Processing} {Methods} for {Regional}-{Scale} {Estimation} of {Manganese}
  {Crust} {Distribution},'' \emph{IEEE Journal of Oceanic Engineering},
  vol.~46, no.~1, pp. 102--114, Jan. 2021.

\bibitem{Umesh2015}
U.~Neettiyath, T.~Sato, M.~Sangekar, A.~Bodenmann, B.~Thornton, T.~Ura, and
  A.~Asada, ``Identification of manganese crusts in {3D} visual reconstructions
  to filter geo-registered acoustic sub-surface measurements,'' in
  \emph{{OCEANS} 2015 - {MTS}/{IEEE} {Washington}}.\hskip 1em plus 0.5em minus
  0.4em\relax IEEE, Oct. 2015, pp. 1--6.

\bibitem{Bodenmann2017}
A.~Bodenmann, B.~Thornton, and T.~Ura, ``Generation of {High}-resolution
  {Three}-dimensional {Reconstructions} of the {Seafloor} in {Color} using a
  {Single} {Camera} and {Structured} {Light},'' \emph{Journal of Field
  Robotics}, vol.~34, no.~5, pp. 833--851, Dec. 2017, arXiv: 10.1.1.91.5767
  ISBN: 9783902661623.

\bibitem{neettiyath_automatic_2022}
U.~Neettiyath, B.~Thornton, H.~Sugimatsu, T.~Sunaga, J.~Sakamoto, and H.~Hino,
  ``Automatic {Detection} of {Buried} {Mn}-crust {Layers} {Using} a
  {Sub}-bottom {Acoustic} {Probe} from {AUV} {Based} {Surveys},'' in
  \emph{Oceans 2022}, Chennai, 2022.

\bibitem{Umesh2017}
U.~Neettiyath, B.~Thornton, M.~Sangekar, K.~Ishii, T.~Sato, A.~Bodenmann, and
  T.~Ura, ``Automatic {Extraction} of {Thickness} {Information} from
  {Sub}-{Surface} {Acoustic} {Measurements} of {Manganese} {Crusts},'' in
  \emph{Oceans 2017 - {Aberdeen}}.\hskip 1em plus 0.5em minus 0.4em\relax IEEE,
  Jun. 2017, pp. 1--7.

\bibitem{Yamada2020}
T.~Yamada, A.~Prügel-Bennett, and B.~Thornton, ``Learning features from
  georeferenced seafloor imagery with location guided autoencoders,''
  \emph{Journal of Field Robotics}, no. October 2019, pp. 1--16, 2020.

\bibitem{Cheng2016}
G.~Cheng, P.~Zhou, and J.~Han, ``Learning {Rotation}-{Invariant}
  {Convolutional} {Neural} {Networks} for {Object} {Detection} in {VHR}
  {Optical} {Remote} {Sensing} {Images},'' \emph{IEEE Transactions on
  Geoscience and Remote Sensing}, vol.~54, no.~12, pp. 7405--7415, Dec. 2016.

\bibitem{Yamada2021}
T.~Yamada and A.~Pr, ``{GeoCLR} : {Georeference} {Contrastive} {Learning} for
  {Efficient} {Seafloor} {Image} {Interpretation},'' \emph{ArXiv}, 2021, arXiv:
  2108.06421v1.

\bibitem{scikit-learn}
F.~Pedregosa, G.~Varoquaux, A.~Gramfort, V.~Michel, B.~Thirion, O.~Grisel,
  M.~Blondel, A.~Müller, J.~Nothman, G.~Louppe, P.~Prettenhofer, R.~Weiss,
  V.~Dubourg, J.~Vanderplas, A.~Passos, D.~Cournapeau, M.~Brucher, M.~Perrot,
  and E.~Duchesnay, ``Scikit-learn: {Machine} {Learning} in {Python},''
  \emph{Journal of Machine Learning Research}, vol.~12, pp. 2825--2830, Jan.
  2011, arXiv: 1201.0490 ISBN: 1532-4435.

\end{thebibliography}

\end{document}